\documentclass{article}
\usepackage{epsfig, graphicx, amsmath, amssymb}

\newcommand{\Dmn}{\mbox{Dmn}}

\newcommand{\aug}{\mbox{aug}}

\newtheorem{theorem}{Theorem}[section]
\newtheorem{corollary}[theorem]{Corollary}

\newtheorem{definition}[theorem]{Definition}
\newtheorem{remark}[theorem]{Remark}
\newtheorem{example}[theorem]{Example}

\begin{document}
\begin{center}
{\bf\large Nominal Association Measures  for Categorical Data}
\end{center}
 
\begin{center}
{ Wenxue Huang, 
Yong Shi,  
 Xiaogang(Steven) Wang\\
 Shantou University, Chinese Academy of Science, York University}
 \end{center}
 
\begin{abstract}
We introduce an intrinsic and informative local-to-global association matrix to measure the proportional association of categories of a variable with another categorical variable.  Towards a proportional prediction, the association matrix determines the expected confusion matrix of a multinomial response variable. The normalization of the diagonal of the matrix gives rise to an association vector, which provides the expected category accuracy lift rate distribution.
 A general scheme of global-to-global association measures associated with flexible weight vectors is
 derived from the diagonal.   A hierarchy of equivalence relations defined by the association matrix and vector is shown. An application to feature selection is presented.
 \end{abstract}


\section{INTRODUCTION}
\label{sect:Intro}
Nominal data are quite common in
scientific and engineering research related to biomedical research, consumer behavior analysis, network analysis and search engine marketing optimization.
When the population is cross-classified and  there is no natural ordering for  observed outcomes,
association analysis as described in Han and Kamber (2006) can be described nominal association measures.
Even if the categorical variables collected in these studies are  ordinal, they are often treated as nominal if the ordering is not of interest or a natural and meaningful metric  is difficult to establish.

When the response variable is multinomial, the classical probabilistic  measure such as odds ratio or relative risk are difficult to use due to the multiple levels in the response variable. Instead, the principle of optimal (conditional mode based) or proportional (conditional Monte-Carlo based) prediction can be  used to construct nonparametric nominal association measures.
For example,  Goodman-Kruskal (1954) and others proposed some local-to-global association measures towards optimal predictions.
The proportional associations between variables are probabilistically and statistically intrinsic. It reflects the probabilistically averaging effects of input on output distributions.
There are quite a few proportional association measures proposed in the literature (cf.~ Lloy (1999)). However,
all these measures focus either on local-to-local (e.~g.~, contingent table analysis) or on global-to-global (e.~g.~, the Goodman-Kruskal $\tau$ by Goodman-Kruskal (1954)  associations between two categorical variables.
 For some statistical inferences,  commonly used local-to-local (category-to-category) or global-to-global (variable-to-variable) association measures can be sufficiently informative.
 When the population is cross-classified, effective local-to-global (category-to-variable) association measures are necessary to provide a more detailed description or evaluation of the intrinsic dependence structure.   For example, these kinds of measures are of fundamental importance in targeted or group specific risk analysis such as  credit risk migration analysis, clinical diagnosis, or public health management.

  In order to gain important insights of the underlying dependence structure, we propose a nominal association matrix to measure the probabilistic proportional associations of every category of a response variable with an explanatory variable.
   We will show that this matrix estimates expected confusion matrix for multinomial response variables
     when a proportional (conditional Monte-Carlo) prediction is employed. Detailed discussions on proportional prediction can be found
     in  Goodman-Kruskal (1954).

The diagonal of the matrix also induces our association vector introduced in this article.  Each component of the association vector represents the proportional  dependence degree of a given category of a response variable with a given explanatory variable.
Furthermore, combined with various sets of admissible weight vectors, the diagonal of the propose matrix provides a general scheme of
generating different global-to-global association measures which embraces the seminal the Goodman-Kruskal $\tau$ (the GK-$\tau$).
  The weighting scheme in the general class of association measure is now explicit when compared with that of the GK-$\tau$. Furthermore, one can now design various association measures according to  different objectives.
 Consequently, scientists are no longer constrained by the limited number of association measures in the literature that might or might not be optimal for their inferential needs.  For example, association measures designed to capture rare events should be different from those aimed at describing the entire population.
More importantly, any prior knowledge or relevant expert opinions can be formally incorporated into this framework.

A hierarchy of equivalence relations induced by the association matrix and vectors are also presented to help understand
 the strengths of proposed association measures.
 This hierarchy is of fundamental interest to the structural analysis for multivariate categorical data.
  We also show that  the association vector, the association matrix and any global association measure determined by a weight vector, particularly the GK-$\tau$, are equivalent only
 when the response variable is binary.

 For high dimensional data, the proposed global association measures are also capable of evaluating the corresponding collective effect of a set of explanatory variables on a given response variable in order to remove redundancy. We prove that a dimensionality reduction for high dimensional categorical data with a response variable is theoretically possible by using the proposed association measures.

This paper is organized as follows. Section 2 introduces  the nominal association matrix.  The proposed association vector    and   global association measures associated with the vector are derived in Section 3.  We   present the hierarchy of equivalence relations induced by the association matrix, association vector, and the GK $\tau$ in section 4.
 Section 5 provides theoretical results  for feature selection in  categorical data.
 We  illustrate the  association matrix and vector by examining two examples in Section 6. Results from simulation studies and real data analysis for feature selection are also presented.  Discussion is  provided in Section 7.  The  Appendix holds the proofs of the theorems or justifications of the assertions in the body.

\section{Association Matrix}
\label{Sect21}

  We propose an association matrix to measure a complete proportional local-to-global association of a categorical response variable with a categorical explanatory variable.

 Let $X$ and $Y$ be categorical variables with domains $\Dmn(X)=\{1,2,\dots, n_X\}$ and $\Dmn(Y)=\{1,2,\dots, n_Y\}$, respectively.
To provide a complete description of proportional local-to-global association, we introduce a proportional association matrix as follows:
\begin{definition}  The association matrix is given by
 \begin{equation}\label{gammaytox}
 \gamma(Y|X):=(\gamma^{st}(Y|X)),
 \end{equation}
where
\[ \gamma^{st}(Y|X):=\frac {E[p(Y=s|X)\;p(Y=t|X)]}{p(Y=s)},\]
 where $s,t=1,2,\cdots, n_Y$.
\end{definition}

The  $(s,t)$-entry $\gamma^{st}(Y|X)$ of the association matrix $\gamma(Y|X)$  is
 the  probability  of assigning or predicting   $Y=t$
  when the truth is in fact $Y=s$ given the information from the
  explanatory variable.
   The key idea is inspired  by the   proportion prediction
  or equivalently the reduction in predictive error employed by  Goodman and Kruskal (1954). However, they only focused on correct predictions and did not differentiate
 various scenarios associated with a wrong prediction.  The association matrix, on the other hand, provides a complete and detailed description in comparison.

This matrix is of both theoretical and practical importance since it can be used to estimate both the first and second type error rate distributions for a proportional prediction.
\begin{remark}\label{rmk:assocMtrx} It can be seen  that the association matrix $\gamma(Y|X)$ has the following properties:
 \begin{enumerate}
    \item[\rm(1)] $\gamma(Y|X)$ is a row stochastic matrix; in general, it is asymmetric.
    \item[\rm(2)] Based on a representative training set, the diagonal entries are exactly the expected accuracy rates of prediction of corresponding categories of the response variable; while the off-diagonal entries of each row are exactly the expected first type error rates of the corresponding category.
    \item[\rm(3)] Based on a representative training set,  the off-diagonal entries of each column are exactly the expected second type error rates of the corresponding category; thus the matrix can be regarded as the expected confusion matrix for a proportional prediction model with a binomial or multi-nominal response variable.
    \item[\rm(4)] $\gamma(Y|X)$ draws a complete local-to-global association picture.  There are two extreme cases: when $Y$ is completely determined by $X$,
      then $\gamma(Y|X)$ is the identity matrix of degree $n_Y$; when $Y$ is independent of $X$, $\gamma^{st}(Y|X)=p(Y=t)$ for all $s$ thus the matrix $\gamma(Y|X)$ is an idempotent with all rows identical to the marginal distribution of $Y$.
    \item[\rm(5)] The proportional association matrices $\gamma(Y|X)$ can be any row stochastic matrix between a rank one idempotent and identity matrices.  (The study of the most detailed local-to-global association measure (the association matrices) can be a big topic.)
 \end{enumerate}
\end{remark}

 Some of its further properties will be presented with context in the subsequent three sections.

 \begin{example} To demonstrate the application of
 the matrix and its properties (2) - (4), we present a case test with a data set of which the values of $(X,Y)$ are determined by Table \ref{Tableone} in Section \ref{Sect4}). In this data set, there is a response variable, $V4$,  from a data set with 120 variables and 24,000 observations.

The response variable has six categories with the following probability distribution:
$$p(Y)=( \;  0.1041,\;    0.3077, \;    0.3060, \;    0.1581, \;    0.1099, \;    0.0143\;).$$
An  explanatory variable $X=V6$ is selected and the association matrix based on the 80\% samples is calculated.
We then use the information   in $X$ and apply the proportional prediction
  to predict the value of the response variable in the remaining 20\% of the original samples.

The association matrix  based on training (left) and the (proportional prediction based) confusion matrix derived from testing  (right) are provided as follows:
\[\left( \begin{array}{cccccc}
    .26   &  .47   &   .15   &  .06   &  .04   &  .01\\
    .05   &  .48   &   .28   &  .11   &  .07   &  .01\\
    .02   &  .36   &  .34    & .15   &  .11   &  .02\\
    .02   &  .32   &  .35    & .17   &  .12    & .02\\
    .02   &  .30   &  .35    & .18   &  .14    & .03\\
    .03   &  .29   &  .33    & .18    & .15    & .03\\
\end{array}\right) VS \left( \begin{array}{cccccc}
    .27  &   .47  &   .16  &   .05 &   .03  &   .01\\
    .05 &    .49  &   .28  &   .10 &     .06 &     .01\\
    .02 &    .36 &    .35 &    .15 &    .10  &   .02\\
    .02  &   .31   &  .36  &   .17 &    .12 &    .03\\
    .02   &  .28 &    .35  &   .17  &   .14 &    .04\\
    .03  &   .27  &   .33  &  .18  &   .15 &    .04\\
\end{array}\right) \]
We see that the confusion matrix by using the 20\% test samples is very close to the association matrix by using the 80\% training samples.
The small difference could be  due to different  sizes of the training and testing sizes and  possible selection bias associated
with pseudo-random numbers generated by the computer.
\end{example}

\section{Association Vector and Global-to-Global Association Degrees}
\label{sec:assVectorAndGlobal}
In this section, we introduce an association vector as an important derivative of the association matrix.  We also construct a class of global-to-global association measures  using  these local-to-global association degrees.   We shall see that the GK $\tau$ belongs to this class and has hierarchical relations with the local-to-global association matrix.   We shall also show that these global-to-global association measures can be applied to variable selection for high dimensional categorical data.

\subsection{Association Vector}

\begin{definition}
 The {\it  association vector}
 $$\Theta^{Y|X}:=(\theta^{Y=1|X},\theta^{Y=2|X},\dots, \theta^{Y=n_Y|X})$$
 is given by
  \begin{equation}\label{thetaysceOnx}
  \theta^{Y=s|X}:=\frac {\gamma^{ss}(Y|X)-p(Y=s)}{1-p(Y=s)},\qquad s=1,2,\dots, n_Y.
  \end{equation}
\end{definition}

 Thus, $\theta^{(Y=s)|X}$ is the normalization of the diagonal entry $\gamma^{ss}(Y|X)$, which is the expected accuracy lift rates of proportional prediction of $Y=s$ using  the information of the explanatory variable $X$ over using only the marginal distribution of $Y$.  It can be verified that
\begin{equation}\label{gammaAndtheta}
\theta^{Y=s|X}=\frac {E[p(Y=s|X)^2]-p(Y=s)^2}{p(Y=s)\;(1-p(Y=s))}.
\end{equation}

 When a proportional predictive model based on  $X$ is deployed, the components of the vector $\Theta^{Y|X}$ are exactly the error reduction (or accuracy lift) rates for proportional prediction over those using the information of $Y$ only.

\subsection{Framework of Global-to-Global Association Measures}

In order to evaluate the overall proportional dependence of a response variable on an explanatory variable,
we define a general class of global-to-global association measures.
 \begin{definition}\label{defnalpha}
 Given a weight vector ${\boldsymbol \alpha}=(\alpha_1, \alpha_2, \dots, \alpha_{n_Y})$ with $\sum_s\alpha_s=1$ and $\alpha_s\geq 0$ for all $s=1,2,\dots, n_Y$, the global association degree is defined as
 $$\tau_{\boldsymbol\alpha}^{Y|X}=\sum_{s=1}^{n_Y}\;\alpha_s\;\theta^{Y=s|X}.$$
 \end{definition}


We call $\tau_{\alpha}^{Y|X}$ the  $\boldsymbol\alpha$-association degree of $Y$ on $X$.  We call a weight vector $\boldsymbol\alpha$ {\it regular} if $\alpha_s> 0$ for all $s=1,2,\dots, n_Y$ when  every single   scenario of $Y$ is considered in the evaluation of the overall
nominal dependence.

The  weight vector provides a practitioner with a useful tool to place a desired emphasis on certain scenarios given different
inferential objectives.  In particular, each component of the association vector can be reproduced by placing $\alpha_s=1$ for a given $s$.
 When dealing with an inference problem, one may choose a set of weights that is relevant and suitable.   Goodman and Kruskal (1954) proposed a general framework to chose optimal weights by the introduction of loss functions. Although their purpose is mainly for gaining understanding of some specific global association measures, it is still applicable to our case. In general, one can first define a loss function
$L(\boldsymbol \alpha)$. To obtain a set of weights, we then  derive the optimal weights from the following minimization problem:
\begin{equation*}
\min\limits_{{\boldsymbol  \alpha}} L({\boldsymbol\alpha})\mbox{ subject to } \sum\limits_{s=1}^{n_Y} \alpha_s=1\mbox{ and } \alpha_s\geq 0, s=1,2,\cdots, n_Y.
\end{equation*}

A certain global-to-global association is then a function of any specified loss function. Consequently, it is objective-dependent as it should be. For example, an objective function for rare events should be different from
 that designed for other general inquiries concerning  the majority of the population.
 For demonstration purposes, we shall only consider three association measures using some commonly used weight functions in the section of examples and simulation study.

\begin{theorem}\label{THM1} Assume $\boldsymbol \alpha$ is a regular weight vector.
\begin{enumerate}
\item[\rm(i)]  $0\leq\theta^{(Y=s)|X}\leq 1$ and $0\leq\tau_{\boldsymbol\alpha}^{Y|X}\leq 1$;
\item[\rm(ii)]  $\theta^{Y=s|X}=0\iff \{Y=s\}$ and $\{X=i\}$ are independent for all $i\in\Dmn(X)$; in particular,
$\tau_{\boldsymbol\alpha}^{Y|X}=0\iff Y$ and $X$ are independent;
\item[\rm(iii)]  $\theta^{Y=s|X}=1\iff p(Y=s|X=i)=1$ or $0$, for all
$i\in \Dmn(X)$; in particular, $\tau_{\boldsymbol\alpha}^{Y|X}=1\iff Y$ is completely determined by $X\iff\tau^{Y|X}=1$;
\end{enumerate}
\end{theorem}
\begin{remark} It follows from the above theorem and Remark \ref{rmk:assocMtrx} that, when $Y$ is completely determined by or independent of $X$, with any regular weight vector $\boldsymbol\alpha$, the association measure $\tau_{\boldsymbol\alpha}^{Y|X}$ completely determines the association vector and matrix.
\end{remark}
We now show that the GK-$\tau$ can be derived from the general association measure.
Recall that the GK $\tau$ (see Goodman-Kruskal (1954)) is given by
\begin{equation}\label{eq3}
\tau^{GK}= \frac{  \sum\limits_{i=1}^{n_Y} \sum\limits_{j=1}^{n_X} p(Y=i; X=j)^2/p(X=j)
- \sum\limits_{i=1}^{n_Y} p(Y=i)^2    } { 1-  \sum\limits_{i=1}^{n_Y} p(Y=i)^2  }.
\end{equation}
It is a also normalized conditional Gini concentration, measuring a proportional global-to-global association.
Detailed discussions can be found in Lloyd  (1999).

We now show that the GK $(\tau)$ can be derived from the proposed framework by using a specific set of weights.

 \begin{corollary}\label{CorGK}
   If the weight vector is assigned as
 \begin{equation}\label{alphaP}
 {\boldsymbol\alpha}^{P}=\frac 1{V_G(Y)}(p(Y=1) - p(Y=1)^2,\dots, p(Y=n_Y) - p(Y=n_Y)^2);
 \end{equation}
 then
\begin{equation}\label{tauEq}
\tau^{GK} = \tau_{{\boldsymbol\alpha}^{P}}^{Y|X}.
\end{equation}
 \end{corollary}

\section{Hierarchy of Equivalence Relations Induced by Association Measures}

We now show the hierarchy of equivalence relations defined by the association matrix, association vector and
a global association degree (in particular, the GK $\tau$).
We denote by $\mathcal C$ the set of explanatory variables and present the five equivalence relations on $\mathcal C$ as follows.

\begin{definition}\label{def3} Let $X_1, X_2\in {\mathcal C}$ and $Y$ a given response variable.  With respect to $Y$, the variables $X_1$
and $X_2$ are
\begin{enumerate}
  \item[\ref{def3}.1] {\it E-1 equivalent},
  if  $\tau^{X_1|X_2}=\tau^{X_2|X_1}=\tau^{Y|X_1}=1$;
  \item[\ref{def3}.2]
             {\it E-2 equivalent}, if  $\tau^{Y|X_1}=1=\tau^{Y|X_2}$;
  \item[\ref{def3}.3]
              {\it E-3 equivalent}, if $\gamma(Y|X_1)= \gamma(Y|X_2);$
  \item[\ref{def3}.4]
             E-4 equivalent, if
             $\Theta^{Y|X_1}=\Theta^{Y|X_2}$;
  \item[\ref{def3}.5]
             E-5 equivalent with respect to a weight vector $\alpha$,
             if  $\tau_\alpha^{Y|X_1}=\tau_\alpha^{Y|X_2}$.
\end{enumerate}
\end{definition}

\begin{theorem}\label{THM2}  All the above defined binary relations E-i,
$i=1,2,3,4,5$, are indeed equivalence relations on the set
$\mathcal{C}$. Furthermore, if $X_1$ and $X_2$ are E-i
equivalent (with respect to $Y$), then
             they are E-(i+1) equivalent (with respect to $Y$), for $i=1, 2, 3, 4$.

If  $Y$ is binary, then for any weight vector $\boldsymbol \alpha$,  with respect to $Y$, the E-$i$
equivalence relations are the same for $i=3,4,5$; and
$\theta^{Y=s|X} = \tau_{{\boldsymbol\alpha}}^{Y|X}, s\in\Dmn(Y)$.
\end{theorem}

This theorem implies that when $Y$ is binary, $\tau_{\alpha}$ (in particular, the GK $\tau$) sufficiently captures the nominal association in this case.

\begin{remark}\label{rmk:forTHM2} If $n_Y>2$, for each $i=1,2,3,4$, there exist $X_1$ and $X_2$, which are E-(i+1)
equivalent (with respect to $Y$), but
             not E-i equivalent (with respect to $Y$).  Counter examples and additional comments are shown in the Appendix.
\end{remark}

 \section{Feature Selection}
\label{subsect31}
We consider feature selection problems with and without response variables.

\subsection{With Response Variable}

In this subsection, we consider the existence of a structural basis for a given data set  $\mathrm{S}$ with  categorical variables $V_1, V_2, \dots, V_n$. with   a response variable $Y$.
We now present the variable selection result for high dimensional data when there is a response variable.

\begin{definition}\label{def5p1} Given a set of  explanatory variables $V_1,\dots, V_n$ and a response variable
$Y$. Let $\alpha$ be a regular weight vector.  A subset $(V_{i_1}, \dots, V_{i_k})\subseteq \mathrm{V}$ is called an $\alpha$-association basis for
$Y$ over $\mathrm{S}$ if
\begin{enumerate}
   \item[TB1.] $\tau_{\boldsymbol \alpha}^{Y|(V_{i_1}, \dots, V_{i_k})}=\tau_{\boldsymbol \alpha}^{Y|(V_1, V_2, \dots, V_n)}$;
   \item[TB2.] for any $V\in \{V_{i_1},\dots, V_{i_k}\}$,\\
           $\tau_{\boldsymbol \alpha}^{Y|(\{V_{i_1},\dots, V_{i_k}\}\setminus\{V\})}<\tau_{\boldsymbol \alpha}^{Y|(V_1, \dots, V_n)}$.
\end{enumerate}
\end{definition}

The following theorem is about the existence, construction and
some properties of the structural bases. The argument in its proof of the inequality (\ref{tag:tauIncrease}) is intrinsic, informative and intriguing.
\begin{theorem}\label{FSTHM}
Let $S$ be a set of  explanatory variables $X_1,\dots, X_n$, $Y$ a response variable, and
$\alpha$ a weight vector.
Then there exists an $\alpha$-association basis $X(i_1,\dots,
i_k):=\{X_{i_1}, X_{i_2}, \dots, X_{i_k}\}$, where $1\leq
i_1<i_2<\dots <i_k\leq n$.
\end{theorem}

\subsection{Without Response Variable}
\label{sect:EBMD}

In the analysis of high dimensional nominal data, the dimensionality reduction is
one the most challenging problem. In this section, we prove the existence of bases for multivariate
nominal data.

 Given a data set
$\mathrm{S}$ with categorical variables $V_1, V_2, \dots, V_n$ and
records of realization such as
 $$R_i:=(x_{i1}, x_{i2}, \dots, x_{in}),\qquad i=1,\dots, m.$$

We consider the following two conditions:
\begin{definition}\label{def4p1} A subset $$\{v_{i_1},
V_{i_2},\dots, V_{i_k}\}\subseteq \mathrm{V}$$ is called an $\alpha$-structural basis for $\mathrm{S}$ for a given regular weight vector $\alpha$ if
\begin{enumerate}
   \item[$\alpha$B1.] for each $V_i\in \mathrm{V}$, $\tau_\alpha^{V_i|(V_{i_1}, V_{i_2},
           \dots, V_{i_k})}=1$;
   \item[$\alpha$B2.] for any $v\in \{V_{i_1},\dots, V_{i_k}\}$,\\
           $\tau_\alpha^{V|(\{V_{i_1},\dots, V_{i_k}\}\setminus\{V\})}<1$.
\end{enumerate}
\end{definition}

We now prove the existence of such a base.
\begin{theorem}\label{thmBMVnoY}
Let $S$ be a data set with variables $(X_1,\dots, X_n)$ with samples representative to distribution but without response variables.  Then
there exists and one can find a structural basis $(X_{i_1},\dots, X_{i_k})$, where $k\leq n$, for the multivariate distribution.
\end{theorem}
  It is worth noting that there can be two or more
$\alpha$-structural bases. But any two such structural bases are E-1
equivalent w.~r.~t.~ the distribution of $S$.  For a high
dimensional (categorical) data set $S$, it can be costly to find a
structure basis $\{V_{i_1},\dots, V_{i_k}\}$ with $k$ smallest. Due
to the following proposition, most of time, $k$ does not need to
be smallest.

\begin{theorem}\label{prop4p8}
Let $S$ be a data set $S$ with variables $V_1,\dots, V_n$, and
$\alpha$ a regular weight vector.

\begin{enumerate}
\item[(1)] Assume that $V(i_1, i_2,\dots, i_k)$ is a subset of
$\mathrm{V}$ in the data set $S$.  If $\{(V_{i_1},\dots,
V_{i_k}\}$ satisfies $\alpha$B1 then $$\tau_\alpha^{V(j_1,\dots,
j_l)|V(i_1,\dots, i_k)}=1$$ for any subset $V(j_1,\dots,
j_l)$ of $\mathrm{V}$.
\item[(2)] If both $V(i_1, i_2,\dots, i_k)$ and $V(j_1, j_2,\dots, j_l)$
are bases for $\mathrm{S}$, then $$|\Dmn(V(i_1,  \dots,
i_k))|=|\Dmn(V(j_1,\dots, j_l))|,$$ which is up bounded by
$$\min(\prod_{s=1}^km_{V_{i_s}},\prod_{t=1}^lm_{V_{j_t}}).$$

\item[(3)] There exists a smallest $k\leq n$ and a subset $B(i_1,\dots,
i_k):=\{V_{i_1}, V_{i_2}, \dots, V_{i_k}\}$, where $1\leq
i_1<i_2<\dots <i_k\leq n$, such that the structure of the whole
$S$ is completely determined by $B(i_1,\dots, i_k)$, i.~e.~, for
any variable $V_i$, and for any $s\in\Dmn(V_i)$, and joint scenario $(V_{i_1}=t_{i_1}, \dots,
V_{i_k}=t_{i_k})$, $p(V_i=s|V_{i_1}=t_{i_1}, \dots,
V_{i_k}=t_{i_k})$ is equal to either $1$ or $0$.
\end{enumerate}
\end{theorem}

\section{DATA ANALYSIS AND SIMULATION STUDY}
\label{Sect4}

\subsection{Examples for Association Vector and Matrix}
\label{subsect41}

To illustrate, we first present results from  two examples including  a credit risk management data set.

\begin{example}\label{example4_1}
We first consider an actual response variable, $V4$,  from a data set with 120 variables and 24,000 observations.
The response variable has six scenarios with the following probability distribution:
$$p(Y)=( \;  0.1041,\;    0.3077, \;    0.3060, \;    0.1581, \;    0.1099, \;    0.0143\;).$$

\rm(i) To demonstrate, we generate another categorical variable  which is independent of the response variable.
 The generated explanatory variable  has six categories.
Since the response and explanatory variables are independent,  every component of the association vector  should also be zero in theory.    The global-to-global $\alpha$-association degree
should be zero for any given  weight vector $\alpha$.
 The estimated association vector is calculated to be
\[
\Theta= (2\times10^{-3},   2\times10^{-3},    3\times10^{-3} ,    10^{-3},   5\times10^{-3},   3\times10^{-3}).
\]
\begin{table}
\caption{\label{Tableone} \small Joint frequency table of the response variable and another variable from a real data set with 24,000 observations.}
\centering
\begin{tabular}{|l|rrrrrr|r|}
\hline $X\setminus Y$&1&2&3&4&5&6&$(X,\cdot)$\\\hline
1&16          &1       &0       &0     &0    &0      &17\\
2&1,199        &1,274    &346     &66    &33   &1       & 2,919\\
3& 640        &2,363    &1,363    &343   &103  &7       & 4,819\\
4& 381        &2,203    &2,646    &949   &402   & 18     &6,599 \\
5&  182       &1,131    &2,038    &1,369  &762   & 55     &5,537\\
6&    79      & 407    &937    & 1,047   &1,286  &206    & 3,962\\
7&    2       &  5      & 14      &20   & 51  & 55     & 147\\\hline
$(\cdot, Y)$& 2,499& 7,384& 7,344&    3,794&   2,637& 342& 24,000\\\hline
\end{tabular}
\end{table}
\vskip 0.1in

To illustrate the idea of the proposed association vector, we consider a response variable, say, V4, with six scenarios, from a real data set with $24,000$ observations (refer to the frequency table below). We
now compute the association vector with a categorical variable which is completely independent of the chosen response variable. Here, the explanatory
variable is generated randomly and independently from the response variable with 6 (can be replaced with any not-too-large integer $N$) categories as well.

The marginal probability distribution of $Y$ is
$$p(Y)=( \;  0.1041,\;    0.3077, \;    0.3060, \;    0.1581, \;    0.1099, \;    0.0143\;).$$
The   association vector calculated is
\[
\Theta= (.0002,   .0002,    .0003 ,   .0001,   .0005,   .0003),
\]
which is very close to a zero vector. As expected, the general association degree will also be very to zero given any weight vector.

We then consider the same response variable as above used and select an actual explanatory variable contained in the same data set.

 Joint frequency is provided in Table \ref{Tableone}
 \[ \tau=.0763. \;\;\;\;\;  \Theta = (.2437,\;.0778,\;.0236,\;.0413,\;.0806,\;.0355).  \]

\rm(ii) Next we  select an actual explanatory variable, $V34$,  contained in the same data set. The joint frequency table is given in  Table \ref{Tableone}.
The GK $\tau$ and the association vector are shown below:
\[ \tau=.0825; \;\;\;\;\;  \Theta = (0.1617, \;   0.0883, \;   0.0418, \;   0.0565, \;   0.1181,  \;  0.0802).  \]
This shows that the selected actual explanatory variable has some demonstrated mild impact on the response variable.
\end{example}

\begin{example}\label{Example4_2}
We now consider a real loan application data set discussed  in Olson and Shi  (2007) and Seppanen {\it et al.} (1999).
This data set has several variables and 650 records.
For simplicity, we are only concerned about the following five (categorical or discretized) variables: {\it On-Time, Age, Income, Credit} and {\it Risk}, where these  variable were categorized as
On-Time=(No (0), Yes (1)); Age=(young, med, sen); Income = (low, mid, hi); Risk =(low, med, hi); Credit = (red, yellow, green).
We consider three situations in which {\it On-Time},  {\it Risk}, and {\it Credit}  are used as the response variable respectively.

1. For response $Y=$ {\it On-Time}, we observe that $p(Y)=(0.1, 0.9)$.  Since $Y$ is binary, by Theorem \ref{THM2}, $\tau_{\alpha}^{Y|X}=\tau^{Y|X}=\theta^{Y=i|X}, i=0,1$.\\
\begin{center}
\begin{tabular}{|c|rrrr|}\hline
$X$&Credit &Risk&Age&Income\\\hline
$\tau^{Y|X}$&.0577&.0486&.0402&.0134\\\hline
\end{tabular}
\\
\end{center}

\vskip 0.1in
2. For the response variable $Y=$ {\it Risk}: $p(Y)=(.4877, .0400, .4723)$, we obtain the following results:\\
\begin{center}
\begin{tabular}{|l|l|l|l|l|}\hline
X &$\tau^{Y|X}$&$\Theta^{Y|X}$&$\gamma(Y|X)$&(X,Y) freq.\\\hline
On-Time & .0432& (.0451,.0002,.0479)&$\left(\begin{smallmatrix}
.5108 &.0407 &.4485 \\ .4959 &.0402 &.4639 \\ .4631 &.0393 &.4976
\end{smallmatrix}\right)$&$\left(\begin{smallmatrix}11 &2 &52 \\ 306 &24 &255 \end{smallmatrix}\right)$\\\hline
Age &.5137 &(.5451,.0018,.5611) &$\left(\begin{smallmatrix}
.7669 &.0437 &.1894 \\ .5324 &.0417 &.4258 \\ .1956 &.0361 &.7684
\end{smallmatrix}\right)$&$\left(\begin{smallmatrix}13 &9 &246 \\ 291 &17 &61 \\ 13 &0 &0\end{smallmatrix}\right)$\\\hline
Income &.0272 &(.0368,.0207,.0185) &$\left(\begin{smallmatrix}
.5065 &.0345 &.459 \\ .4206 &.0599 &.5195 \\ .4739 &.044 &.4821
\end{smallmatrix}\right)$&$\left(\begin{smallmatrix}19 &8 &45 \\ 211 &17 &209 \\ 87 &1 &53 \end{smallmatrix}\right)$\\\hline
Credit &.0009 & (.0006,.0008,.0012) &$\left(\begin{smallmatrix}
.488 &.0401 &.4719 \\ .4892 &.0408 &.4700 \\ .4872 &.0398 &.4729
\end{smallmatrix}\right)$&$\left(\begin{smallmatrix}35 &2 &40 \\ 98 &9 &93 \\ 184 &15 &174\end{smallmatrix}\right)$\\\hline
\end{tabular}
\end{center}

\vskip 0.1in
3. For the response variable $Y=$ Credit: $p(\cdot, Y)=(.1185, .3077, .5738)$, we obtain the following results:\\

\begin{center}
\begin{tabular}{|l|l|l|l|l|}\hline
X &$\tau^{Y|X}$&$\Theta^{Y|X}$&$\gamma(Y|X)$&(X,Y) freq.\\\hline
On-Time & .0319&(.0322, .0123, .0488)&$\left( \begin{smallmatrix}
.1468 &.3328 &.5204 \\
.1281 &.3162 &.5556 \\
.1074 &.2979 &.5946
\end{smallmatrix}\right)$&$\left( \begin{smallmatrix}19 &30 &16 \\
58 &170 &357\end{smallmatrix}\right)$\\\hline
Age &.0035 &(.0099, .0028, .0014)&$\left( \begin{smallmatrix}
.1272 &.3023 &.5705 \\
.1164 &.3096 &.5740 \\
.1178 &.3078 &.5744
\end{smallmatrix}\right)$&$\left( \begin{smallmatrix}40 &80 &148 \\
34 &118 &217 \\
3 &2 &8\end{smallmatrix}\right)$\\\hline
Income &.001 &(.0007, .0006, .0016)&$\left( \begin{smallmatrix}
.1191 &.3085 &.5724 \\
.1188 &.3081 &.5731 \\
.1182 &.3073 &.5745
\end{smallmatrix}\right)$&$\left( \begin{smallmatrix}7 &20 &45 \\
54 &137 &246 \\
16 &43 &82 \end{smallmatrix}\right)$\\\hline
 Risk &.0005 &(.0016,.0003,.0002) &
$\left( \begin{smallmatrix}
.1199 &.3069 &.5733 \\
.1181 &.3079 &.5739 \\
.1183 &.3077 &.5739
\end{smallmatrix}\right)$ & $\left( \begin{smallmatrix}
35 &98 &184 \\
2 &9 &15 \\
40 &93 &174\end{smallmatrix}\right)$\\\hline
\end{tabular}
\end{center}
\vskip 0.1in

The variable {\it Risk} was generated by a seemingly subjective discretization on the ratios of debt over asset and set to reflect the degree of risk of the loan or borrower.  Calculations have shown   that {\it Risk} and {\it Credit} are  almost independent of each other.  Moreover, the variable
    {\it On-Time} is quite lowly associated with each of the two variables.
\end{example}

 Based on the analysis results, we believe that there are two possibilities. Either the credit scoring or the risk assigned is in
poor quality; or even  both are in poor quality. In this case, the continuous variable should be properly rediscretized.  Secondly, the existing categorized Risk is a conventional yet subjective classification of the debt-over-asset ratios.  We can see that this classification is almost irrelevant  for the loan risk management.  There is a comprehensive account of the credit migrations and rating based modeling of credit risk in Trueck and Rachev (2009).  It appears that our association matrix may not only provide a direct estimate of the credit migration matrix but also improve the quality of the credit rating by means of the association measure based discretization of the involved continuous explanatory variables.

\subsubsection{Simulation Study}
\label{subsect43}
Consider a simulated  data set motivated by  a real situation. Suppose that a disease, say, flu, is the concern. Assume
 that there are two types of flues, regular and H1N1 flu.  The response variable takes the value 0 representing the absence of any flu and
 takes values $ 1, 2$ for regular and H1N1 flu respectively.

 Suppose that there are  two types of test that are available. We assume that none of these two  is accurate enough to be conclusive.
 However, we assume that they can be very useful when the results are combined in the sense to be made precise in the following.

For simplicity, these two tests  are assumed to be independent and $P(X_2=1)=P(X_3=1)=1/4$.
The joint distribution of $(Y, X1, X2)$ is given by the following table:
\vskip 0.05in
\begin{center}
{\small
\begin{tabular}{|ccccc|}
\hline
$(X_1, X_2)$ &   $P(X1; X2)$ & $ P(Y=0|X_1, X_2)$ & $P(Y=1|X_1; X_2)$ & $P(Y=2|X_1; X_2)$  \\
\hline
(0, 0) & 9/16 &   95\% & 5\% &  0\%\\
(0, 1) & 3/16 &    30\% & 70\% & 0\%\\
(1, 0) & 3/16 &    50\% & 50\% & 0\%\\
(1, 1)&  1/16 &    0\% & 5\% & 95\%\\
\hline
\end{tabular}
}
\end{center}
\vskip 0.05in

Based on $X_1$ and $X_2$, we also generate  two {\it less informative} variables, $R_3$ and $R_4$  such that
 $P(R_3=1|X_1=1)=P(R_4=1|X_2=1)=0.90$ and
$P(R_3=0|X_1=1)=P(R_4=0|X_2=1)=0.10$.

These two newly created variables $R_3$ and $R_4$ are indeed redundant if
 $X_1$ or $X_2$ are selected in the analysis.
 Furthermore, we generate another variables based on the joint distribution of
$X_1$ and $X_2$. To be more specific, we have $S_5= I(X_2=1)*I(X_3=1)*Z$ where $Z$ is an independent Bernoulli random variable with
$p=0.8$. The marginal distribution of $Y$ is given by $P(Y=0)=0.6875$, $P(Y=1)=0.2531$ and $P(Y=2)=0.0594$.

Consequently, there are only two effective explanatory variables, $X_1$ and $X_2$, since the rest are all derived from
  based on the joint or marginal distributions of $X_1$ and $X_2$. Intuitively, the derived random variables will suffer certain information loss. However, both $X_1$ and $X_2$ must be used simultaneously for an effective prediction of the response variable
  $Y$ while $S_5$ is more indicative than using $X_1$ or $X_2$ alone.

\begin{table}
\caption{ \label{tableSIM} Measure of associations using different weighting schemes.}
\centering
\begin{tabular}{|l|ccccc|}
\hline
 & $X_1$ & $X_2$ & $R_3$ & $R_4$ & $S_5$   \\
\hline
 $\tau$(GK) &   0.2382 &   0.1010 &   0.2060  &  0.0878 &   0.1511         \\
  $\tau$(ew) &   0.2221&    0.1206&    0.1923 &   0.1050 &   0.2943     \\
   $\tau$(ipw) &0.1900    &0.1597   & 0.1648  &  0.1393  &  0.5806     \\
 \hline
 \hline
\hline
 & $X_1+R_4$  & $X_2+R_3+S_5$   & $X_1+R_4+S_5$ & $X_1+X_2$ & ALL   \\
\hline
 $\tau$(GK)  &  0.4627   & 0.4669 &   0.4823 &    0.5018 &   0.5018\\
  $\tau$(ew) & 0.5570    & 0.5731  &  0.5884 &   0.6078  &  0.6078\\
   $\tau$(ipw)   &  0.7372 &   0.7940 &   0.8004 &   0.8198 &   0.8199\\
 \hline
\end{tabular}
\end{table}

For comparison purposes, we consider three weighting schemes:
\begin{eqnarray}
{\bf \alpha}_k^{GK} &=& p(Y=k)(1-p(Y=k))/[\sum\limits_{j=1}^{n_Y} p(Y=j)(1-p(Y=j)];\\
{\bf \alpha}_k^{ew} &=& 1/n_Y;\\
{\bf \alpha}_k^{ipw} &=& 1/p(Y=k)/[ \sum\limits_{j=1}^{n_Y} 1/p(Y=j) ];
\end{eqnarray}
assuming that $p(Y=k)>0$ for any $k$.
 If this is not the case, a proper re-coding for the categories of
$Y$ will eliminate this trivial case.

It is clear that the first weighting scheme is precisely the one
coinciding with the GK $\tau$. The second one is a na\"{i}ve equal weighting scheme which is independent of any
probability distribution. The third one is the weighting scheme using the inverse of the individual probabilities.
This is reasonable when rare but severe events are of major concern. The corresponding associations measures are then called
 $\tau(GK)$, $\tau(ew)$, and $\tau(ipw)$ respectively.

 By simulating 100,000 observations, the association measures for all five explanatory variables  are shown in Table 2 and Figure 1.
  The   variables $R_3$ and $R_4$, which are   less informative and redundant with resect to $X_1$ and $X_2$,  assume smaller association values than their parent variable, $X_1$ and $X_2$, respectively, for all three association measures.
  The inverse probability weighting scheme provides the highest association value for combination of explanatory variables.
 Furthermore, it can be seen that a dimensionality reduction is achieved by using the combination of $X_1, X_2$ by correctly
 eliminating other redundant variables. We also also added many irrelevant explanatory variables and the result remains the same.
 This is consistent with intuition that the inverse probability weighting is more suitable for capturing rare events than
 the GK $\tau$ which is not sufficiently designed or sensitive to inference of this nature.

We also set the sample size to be 500 and number of iterations for the bootstrap to be 1,000. We study the percentage of association reduction using only $X_1$ and $X_2$ when compared with that from using all five variables. For one particular realization followed by a bootstrap procedure using a stratified sampling scheme, the bootstrapped confidence intervals for the three weighting schemes are $: (95.66\%,   99.88\%),   (98.53\%,  99.94\%), (96.28\%,  99.96\%)$ respectively. The corresponding means are 98.33\%, 99.43\% and 98.97\% respectively. It can be seen that the association using only two most important variables only results in some insignificant information loss.
It takes 205 seconds to calculate the 1,000 bootstrap iterations sampling from 500 observations using
MATLAB.7.10  on  a computer desktop with Intel(R)Core(TM)i7 CPU 870 2.93GHz with 6GB RAM on a 64-bit operation system.

\section{CONCLUDING REMARKS}
\label{Sect5}

  The proposed association matrix with its derived association vector provides new
   additions to existing proportional association measures for analyzing categorical data.
    It provides very detailed description of local-to-global association structure that can not be captured by any global-to-global measure when the response variable is non-binary. The matrix given by training samples gives the expected  confusion matrix for a proportional prediction being deployed.

    The association vector and global-to-global association measures offer local and global insights into nominal proportional associations and are also applicable  to  ordinal categorical data.  The proposed framework provides a general method to produce various  association measures for different inferential purposes.  The presented framework can also be applied  to high dimensional contingency tables in national surveys.

      The introduction of association matrix opens the door to further studies of proportional association structure and inference for categorical data.  Our future works include applying the proposed measure to text data and data for biomarkers.

\section*{APPENDIX}

{\bf Proof of Theorem \ref{THM2}}:
 Observe that if $X_1$ and $X_2$ are E-2 equivalent with respect to $Y$, then
$\gamma(Y|X_i)=I_{n_Y}$, the identity matrix of degree $n_Y$, $i=1,2$.

One checks,
\begin{equation}\label{gammaInprob}
\gamma^{st}(Y|X)=\sum_{i\in\Dmn(X)}\frac {p(X=i,Y=s)
 p(X=i, Y=t)}{p(X=i)\;p(Y=s)}.
 \end{equation}
Thus we have
\begin{equation}\label{gammaAndtheta}
\gamma^{ss}(Y|X)=(1-p(Y=s))\;\theta^{Y=s|X}\;+\;p(Y=s).
\end{equation}
Notice that we have
\begin{equation*}
\tau_\alpha^{Y|X}=\sum_{s\in\Dmn(Y)}\alpha_s\theta^{Y=s|X}
\end{equation*}

Now assume $\Dmn(Y)=\{1,2\}$.  It is routine to check that
$$\theta^{Y=s|X}=\tau_{\boldsymbol\alpha}^{Y|X}$$
 for all $s\in\Dmn(Y)$ and any weight vector
$\alpha$.
\begin{align*}
\gamma^{11}(Y|X)&=p(Y=1)+\frac {V_G(Y)\;\tau_{\alpha}^{Y|X}}{2 p(Y=1)},\\
\gamma^{22}(Y|X)&=p(Y=2)+\frac {V_G(Y)\;\tau_{\alpha}^{Y|X}}{2 p(Y=2)};
\end{align*}
by the above argument, we have $\gamma^{11}(Y|X)=\gamma^{22}(Y|X)$.
Notice that $\gamma(y|x)$ is a two-by-two row stochastic matrix.  So the matrix is uniquely determined by $\tau_{\alpha}^{Y|x}$.
$\square$

\parindent 0pt
{\bf Counter examples for Remark \ref{rmk:forTHM2}}.

The E-i equivalence relations depend on the choice of response variable $Y$.
We now show by counter-examples that E-i equivalence is strictly stronger than E-(i+1) equivalence.

1. \ref{def3}.2 $\nRightarrow$ \ref{def3}.1.We are given categorical variables $X_1$, $X_2$ and
$Y$ in a given data set $S$.\\
\begin{enumerate}
  \item[a.]
 Consider the following joint distribution:\\
     \begin{tabular}{|l|cccc|}
    \hline
            $Y$ &  1 & 0  &  0 & 1 \\
   \hline
          $X_1$ &  1 & 2  &  3 & 4 \\
          $X_2$ &  2 & 3  &  1 & 2 \\ \hline
     probability &  2/7 &2/7  & 2/7 &1/7 \\   \hline
   \end{tabular}
   \vskip .1in
 we see that $X_1, X_2, Y$ satisfy \ref{def3}.2 but not
   \ref{def3}.1.

2. To see \ref{def3}.3 $\nRightarrow$ \ref{def3}.2, we first notice
   that
   if $X_1, X_2, Y$ satisfy \ref{def3}.2. We then have
   $\gamma^{st}(Y|X_1)=\delta_{st}=\gamma^{st}(Y|X_2)$
     for all $s, t\in\Dmn(Y)$.  There exist $X_1, X_2,
             Y$ satisfying \ref{def3}.3 but not \ref{def3}.2,
             i.~e.~, $(\gamma^{st}(Y|X_1))=(\gamma^{st}(Y|X_2))\neq (\delta_{st})$.

3. To see \ref{def3}.4 $\nRightarrow$ \ref{def3}.3, we
consider the following joint distribution.\\
\vskip .1in
  \begin{tabular}{|l|cccccc|}
    \hline
       $Y$ &  1 & 2  &  2 & 4  & 3 &  4 \\
   \hline
     $X_1$ &  1 & 1  &  2 & 2  & 3 &  3 \\
     $X_2$ &  1 & 3  &  2 & 3  & 1 &  2 \\ \hline
probability &  1/6 &1/6  &1/6 &1/6  &1/6 &1/6\\
     \hline
   \end{tabular}
   \vskip .1in

Then $\gamma^{qq}(Y|X_i)=\frac 12$, for all $q=1,2,3,4$ and $i=1,2$,
 while $\gamma^{12}(Y|X_1)=\frac 12\neq 0 =\gamma^{12}(X_2)$.

Remark 4. To see \ref{def3}.5 $\nRightarrow$ \ref{def3}.4, we
consider the following joint distribution.\\
  \vskip .1in
  \begin{tabular}{|l|cccccccc|}
    \hline
       $Y$ &  1 &  1 & 2  & 3 &  1 & 2 & 3 & 2\\
   \hline
     $X_1$ &  1 &  1 & 2  & 3 &  4 & 1 & 1 & 4\\
     $X_2$ &  2 &  1 & 1  & 1 &  4 & 1 & 3 & 4\\ \hline
     probability
           &  1/10 &2/10  &1/10 &1/10  &1/10 &2/10&1/10 &1/10 \\
     \hline
   \end{tabular}
   \vskip .1in
Then $\tau^{Y|X_1}=\tau^{Y|X_2}=\frac 9{25}$,
  $(\theta^{Y=1|X_1}, \theta^{Y=2|X_1}, \theta^{Y=3|X_1})=(\frac 16,
\frac {17}{72}, \frac {23}{48})$, \[
(\theta^{Y=1|X_2},
\theta^{Y=2|X_2}, \theta^{Y=3|X_2})=( \frac {17}{72}, \frac 16,
\frac {23}{48}).\]

Observation 1. If we replace E-$2$ equivalence with E-$2'$, where E-$2'$ is defined as ``$X_1$ and $X_2$ are
  E-$2'$ equivalent if $\tau^{X_1|X_2}=1=\tau^{X_2|X_1}$", then
  the stronger-to-weaker chain that
  $$\text{E-$1$ $\implies$ E-$2'$ $\implies$ E-$3$ $\implies$ E-$4$ $\implies$ E-$5$
  (but not vice versa)}$$
   still holds.

    To see that E-$2'$ implies E-$3$, we note that
   that since $\tau^{X_1|X_2}=1=\tau^{X_2|X_1}$. We then have that
   $|\Dmn(X_1)|=|\Dmn(X_2)|$ and for any event $X_1=i$ there is a unique
   $X_2=j$ such that $p(X_2=j|X_1=i)=1$ and vice versa.  Assume
   that $\Dmn(X_1)=\{i_1,\dots, i_k\}$.  Then
   $\Dmn(X_2)=\{j_1,\dots,j_k\}$ and we may and shall assume that
   $$p(X_2=j_q|X_1=i_q)=1=p(X_i=i_q|X_2=j_q),\qquad q=1,\dots,
   k.$$
   Thus
   \begin{align*}
   \gamma^{st}(Y|X_1)&=\sum_{q=1}^k\frac {p(X_1=i_q,
   Y=s)\;p(X_1=i_q, Y=t)}{p(X_1=i_q)\;p(Y=s)}\\
                     &=\sum_{q=1}^k\frac {p(X_2=j_q,
   Y=s)\;p(X_2=j_q, Y=t)}{p(X_2=j_q)\;p(Y=s)}\\
                     &=\gamma^{st}(Y|X_2).
  \end{align*}
  Thus $X_1$ is E-$3$ equivalent to $X_2$.
  It is easy to see in general E-$3$ equivalence does not imply E-$2'$
  equivalence.
\end{enumerate}

Observation 2. E-1 equivalence condition can be replaced with
  ``if  $\tau_\alpha^{X_1|X_2}=\tau_\alpha^{X_2|X_1}=\tau_\alpha^{Y|X_1}=1$
  for a regular weight vector $\alpha$".

 Actually, $\tau_\alpha^{X_1|X_2}=\tau_\alpha^{X_2|X_1}=\tau_\alpha^{Y|X_1}=1$ if and only if
  $\tau^{X_1|X_2}=\tau^{X_2|X_1}=\tau^{Y|X_1}=1$.
Similarly, E-2 equivalence condition can be replaced with ``if  $\tau_\alpha^{Y|X_1}=1=\tau_\alpha^{Y|X_2}$
for a regular weight vector $\alpha$".

In practice, E-1 equivalence is usually not concerned due to its extreme nature.
Two random variables
 $X_1$ and $X_2$
are E-2 equivalent w.~r.~t.~ $Y$ if and only if $Y$ is completely
determined by $X_1$ or $X_2$; and in this case, there exist hard partitions
$\mbox{Par}(X_i):=\{X_i^s| s\in\Dmn(Y)\}$ of $\Dmn(X_i)$, $i=1, 2$, where
each $X_i^s$ consists of some scenarios of $X_i$, such that
$$\{(a_1, a_2)|a_1\in\Dmn(X_1^s), a_2\in X_2^t\}=\emptyset\mbox{
whenever }s\neq t;$$ while $\{(a_1, a_2)|a_1\in\Dmn(X_1^s), a_2\in
\Dmn(X_2^s)\}\neq\emptyset$ for all $s\in\Dmn(Y)$.
This is because if $a_1\in f_1^{-1}(p)$ then
there exists $a_2\in\Dmn(X_2)$ such that $(a_1, a_2, p)\in
X_1\times X_2\times Y$.  Thus $a_2\in f_2^{-1}(p)$. Hence $S(p,
p)\neq \emptyset$.  By the arbitrariness of $a_1$ in
$f_1^{-1}(p)$, we see that $S(p, q)=\emptyset$, when $p\neq q$.
The above argument also shows that $X_1\times X_2=\cup_{p\in\Dmn(Y)}S(p, p)$.

\parindent 0pt
 {\bf Proof of Theorem \ref{THM1}:}
 One checks that
\begin{align}\label{EqA}
 \theta^{(Y=s)|X} =&\frac {\sum_{j\in\Dmn(X)}\frac
 {p(X=j,Y=s)^2}{p(X=j)}-p(Y=s)^2}{p(Y=s)(1-p(Y=s))} \notag\\
    =&\sum_j \frac { \big (p(X=j, Y=s)-p(X=j)p(Y=s)\big )^2}
    {p(X=j)\;p(Y=s)(1-p(Y=s))}\geq 0,
\end{align}
and
 \begin{align}\label{EqB}
 E[p(Y=s|X)^2]&=\sum_j
 p(Y=s|X=j)^2\;p(X=j)\notag\\
 &\leq\sum_j p(Y=s|X=j)p(X=j)=p(Y=s).
 \end{align}
 Thus $\theta^{(Y=s)|X}\leq 1$.
The rest of \rm(i) to \rm(iii) follows from (\ref{EqA}) and (\ref{EqB}). $\square$
\vskip 0.1in

{\bf Proof of Corollary \ref{CorGK}:}
Let us consider
\begin{eqnarray*}
RHS &=& \frac{1}{V_G(Y)} \sum\limits_{i=1}^{n_Y}
 p(Y=i) (1 - p(Y=i)) \;\;\theta^{Y=i|X}\\
 &=& \frac{1}{V_G(Y)} \sum\limits_{i=1}^{n_Y}
 \left ( E[ p(Y=i)|X)^2] - p(Y=i)^2   \right )\\
 &=& \frac{1}{V_G(Y)}
  \left (  \sum\limits_{i=1}^{n_Y} \sum\limits_{j=1}^{n_X} p(Y=i; X=j)^2/p(X=j)  -E p(Y)   \right ).
\end{eqnarray*}
This completes the proof. $\square$

\vskip 0.2in

{\bf Proof of Theorem \ref{FSTHM}}:

First of all, we prove that the association is non-decreasing when a new variable is added.  (We remind the reader that this property is no surprise, while its verification is quite informative and intriguing.)  We may and shall assume
that the weight vector $\alpha$ is regular.
For any two variables, say, $X_1,X_2$, we want to prove that $\tau_\alpha^{Y|(X_1,X_2)}\geq \tau_\alpha^{Y|X_1}$; the equality holds if and only if
$p(Y=s|X_1=i, X_2=j)=p(Y=s|X_1=i)$ for all $s\in\Dmn(Y), i\in\Dmn(X_1), j\in\Dmn(X_2)$.

It suffices to show that for each $s\in\Dmn(Y)$, $\theta^{Y=s|(X_1,X_2)}\geq \theta^{Y=s|X_1}$.  Since
$\theta^{Y=s|X}=\frac {\gamma^{ss}(Y|X)-p(Y=s)}{1-p(Y=s)}$, we need only to prove
that for each $s\in \Dmn(Y)$, $p(Y=s)\gamma^{ss}(Y|(X_1, X_2)\geq p(Y=s)\gamma^{ss}(Y|X_1)$, that is,
$$\sum_{i,j}\frac {p(X_1=i, X_2=j, Y=s)^2}{p(X_1=i, X_2=j)}\geq\sum_i\frac {p(X_1=i, Y=s)^2}{p(X_1=i)}.$$  Indeed, we have

\begin{align}
&p(Y=s)(\gamma^{ss}(Y|X_1, X_2)-\gamma^{ss}(Y|X_1))\notag\\
&=\sum_{i,j}\frac{p(X_1=i, X_2=j, Y=s)^2}{p(X_1=i, X_2=j)}- \sum_i\frac{p(X_1=i, Y=s)^2}{p(X_1=i)}\notag\\
&=\sum_{i,j}\frac{(p(X_1=i, X_2=j, Y=s)p(X_1=i)-p(X_1=i, Y=s)p(X_1=i, X_2=j))^2}{p(X_1=i, X_2=j)p(X_1=i)^2}\geq 0.\notag\\
\label{tag:tauIncrease}
\end{align}

The rest follows from the above last inequality.

Now we go to the feature selection process.
Let
\[ D_1:=\{X_h|\tau_\alpha^{Y|X_h}=\max_{1\leq j\leq n}\tau_\alpha^{Y|X_j}\};\]
identify
\[ i_1=\min \aug\{X_k\in D_1||\Dmn(X_k)|=\min_{X_h\in D_1}|\Dmn(X_h)\}.\]
Identify
\[ i_2=\min \aug\{X_h|\tau_\alpha^{Y|(X_{i_1}, X_h)}=\max_{1\leq j\leq n}\tau_\alpha^{Y|(X_{i_1},X_j)}\}.\]
If $\tau_\alpha^{Y|(X_{i_1}, X_{i_2})}=\tau_\alpha^{Y|X_{i_1}}$, then stop the forwarding selection process; otherwise continue in this fashion, until
$(k+1)$st step for which
$$\tau_\alpha^{Y|(X_{i_1}, X_{i_2}, \dots, X_{i_k},X_{i_{k+1}})}=\tau_\alpha^{Y|(X_{i_1},\dots, X_{i_k})}$$
for any $i_{k+1}\in\{1,\dots, n\}\setminus \{i_1,\dots, i_k\}$.

We then delete all potential redundant $X_{i_j}$s among $X_{i_1}, X_{i_2}, \dots, X_{i_k}$, here a variable $X_{i_j}$ is redundant if
$$\tau_\alpha^{Y|(X_{i_1}, X_{i_2}, \dots, X_{i_{j-1}},X_{i_{j+1}}, \dots, X_k)}=\tau_\alpha^{Y|(X_{i_1},\dots, X_{i_k})}.$$
After these backward steps, the remaining $X_{i_h}$s form an $\alpha$-association basis.
$\square$

\noindent
{\bf Proof of Theorem \ref{thmBMVnoY}}:

 (1). We prove that, for any variables $X, Y$ in $S$,
 \begin{equation}\label{Epxy}
 Ep(X)\geq Ep(X,Y)\geq \frac 1{|Dmn(X,Y)|};
 \end{equation}
 the first quality holds if and only if $\theta^{Y=s|X}=1$ for all $s\in Y$; the second equality holds if and only if $(X,Y)$ is uniformly distributed.

 Indeed, we have that
 $$Ep(X, Y)=\sum_{i,s}p(X=i, Y=s)^2\leq\sum_{i,s}p(X=i)p(X=i,Y=s)=Ep(X);$$
 the equality holds if and only if whenever $p(X=i, Y=s)\neq 0$, the equality $p(X=i, Y=s)=p(X=i)$ holds, which amounts to
 $\sum_i\frac {p(X=i, Y=s)^2}{p(X=i)}=1$ for each $s\in\Dmn(Y)$, that is, $\theta^{Y=s|X}=1$ for each $s\in\Dmn(Y)$.

 (2).  It follows from (1) that $$Ep(X_{j_1},\dots, X_{j_m})\geq Ep(X_1,\dots, X_n)$$
 for any variable subset $\{X_{j_1},\dots, X_{j_m}\}$ of $\{X_1,\dots, X_n\}$;
 and the equality holds if and only if $\{X_{j_1},\dots, X_{j_m}\}$ contains an $\alpha$-structural basis for the joint distribution $(X_1,\dots, X_n)$.
Thus if
\begin{equation}\label{maxepx}
Ep(X_{j_1},\dots, X_{j_m})= Ep(X_1,\dots, X_n),
\end{equation}
 then for any $X_i$ we have
$\theta^{X=s|(X_{j_1},\dots, X_{j_m})}=1$ for each $s\in\Dmn(X)$.
Notice that
 $$Ep(X_1,\dots, X_n)\geq \frac 1{|\Dmn(X_1,\dots, X_n)|};$$
 the equality holds when and only when the joint distribution $(X_1,\dots, X_n)$ restricted to $S$ is uniform.

(3).  If $n$ is not large, one may find a structural basis by identifying $\Dmn(X_1,\dots, X_n)$ first and then following a backward redundance-removing process based on (\ref{maxepx}).  If $n$ is large (i.~e.~, a high dimensional case), the identification of  $\Dmn(X_1,\dots, X_n)$ is difficult and backward process is not feasible.  In this case, we will go with a forward selection process to construct a base.

 (4). We now start our construction by forward selection. Pick a variable $X_{i_1}$ with
  $$i_1\in\aug\{X_{i_j}|Ep(X_{i_j})=\min_{1\leq h\leq n}Ep(X_h).$$
  Then pick a variable $X_{i_2}$ with
  $$i_2\in\aug\{X_{i_j}|Ep(X_{i_1},X_{i_j})=\min_{1\leq h\leq n}Ep(X_{i_1}, X_h).$$
Continue this forward selection process until
$$Ep(X_{i_1}, \dots, X_{i_k})= \min_{1\leq h\leq n}Ep(X_{i_1}, \dots, X_{i_k}, X_h).$$
Then remove all potential redundant variables among $X_{i_1}, \dots, X_{i_k}$, where a variable $X_{i_j}$ is redundant if
$$Ep(X_{i_1}, \dots, X_{i_{j-1}}, X_{i_{j+1}}, \dots, X_{i_k})=Ep(X_{i_1}, \dots, X_{i_k}).$$
We assume without loss of generality that there is no redundant variables among $X_{i_1}, \dots, X_{i_k}$.
We claim that $X_{i_1}, \dots, X_{i_k}$ form a structural basis for the distribution.
$\square$

\noindent
{\bf Proof of Theorem \ref{prop4p8}}:

(1).  Due to Theorem \ref{THM1}, we may and shall take $\alpha=\alpha^P$.  Assume in $S$ that $V(i_1,\dots, i_k)|_S\subseteq
\mathrm{V}|_S$ satisfies $\alpha$B1, and that $V(j_1, \dots, j_l)$ is a subset of $\mathrm{V}$. Since the condition $\alpha$B1 is
satisfied, we have many-to-one maps $$\phi_t: \Dmn(\{V_{i_1},
\dots, V_{i_k}\})\to\Dmn(V_{j_t}),\qquad 1\leq t\leq l,$$ which is
defined by $$\phi_t(x_{i_1}, \dots, x_{i_k})=x_{j_t}\text{ if }
p(V_{j_t}=x_{j_t}|V_{i_s}=x_{i_s}, \quad 1\leq s\leq k)\neq 0.$$
These maps are well-defined, for,
$p(V_{j_t}=x_{j_t}|V_{i_s}=x_{i_s}, \quad 1\leq s\leq k)\neq 0$ if
and only if $p(V_{j_t}=x_{j_t}|V_{i_s}=x_{i_s}, \quad 1\leq s\leq
k)=1$.

Define a map
$$\psi:=\Dmn(V_{j_1})\times\Dmn(V_{j_2})\times\dots\times
\Dmn(V_{j_l})\to \Dmn(V(j_1, j_2,\dots, j_l))\cup\emptyset$$ by
 \begin{align*}
   &\psi(x_{j_1}, x_{j_2},\dots, x_{j_l})=(x_{j_1}, x_{j_2},\dots, x_{j_l})\\
   &\qquad\qquad\text{if }p(V_{j_1}=x_{j_1},V_{j_2}=x_{j_2},\dots, V_{j_l}=x_{j_l})\neq 0;\\
   &\psi(x_{j_1}, x_{j_2},\dots, x_{j_l})=\emptyset\\
   &\qquad\qquad\text{if }p(V_{j_1}=x_{j_1},V_{j_2}=x_{j_2},\dots, V_{j_l}=x_{j_l})= 0.
 \end{align*}
Then $$\chi:=\psi\circ (\phi_1,\phi_2,\dots,\phi_l): \Dmn(V_{i_1},
\dots, V_{i_k}) \to\Dmn(V_{j_1},  \dots, V_{j_l})=\Dmn(V_{j_1},
\dots, V_{j_l})\cup\emptyset$$ is a many-to-one map (i.e., a
deterministic function) which satisfies that
 \begin{align*}
   &\chi(x_{i_1},\dots, x_{i_k})=(x_{j_1},\dots, x_{j_l}),\\
   &\qquad\qquad\text{if }p(V_{j_1}=x_{j_1},\dots, V_{j_l}=x_{j_l}|
      V_{i_1}=x_{i_1},\dots, V_{i_k}=x_{i_k})\neq 0;\\
   &\chi(x_{i_1},\dots, x_{i_k})=\emptyset,\\
   &\qquad\qquad\text{if }p(V_{j_1}=x_{j_1},\dots, V_{j_l}=x_{j_l}|
       V_{i_1}=x_{i_1},\dots, V_{i_k}=x_{i_k})=0.
 \end{align*}
If $(x_{j_1},\dots, x_{j_l})\in\Dmn(V(j_1,\dots,j_l))$ then
$\exists (x_{i_1},\dots, x_{i_p})\in \Dmn(V(i_1, \dots, i_k))$ such that $$A:=p(V_{j_1}=x_{j_1}, \dots,
V_{j_l}=x_{j_l}|V_{i_1}=x_{i_1}, \dots, V_{i_k}=x_{i_k})\neq 0.$$
Thus $$P(r):=p(V_{j_r}=x_{j_r}|V_{i_1}=x_{i_1},\dots,
V_{i_k}=x_{i_k})\neq 0, \qquad r=1,\dots, k$$
 and then by $\alpha$B1,
$$P(r)=1, \qquad r=1,\dots, k.$$
 If $A<1$, then there exists an $r_0$ such that $$p(V_{j_{r_0}}\neq x_{j_{r_0}}|V_{i_1}=x_{i_1},\dots,
V_{i_k}=x_{i_k})\neq 0,$$
 implying $P(r_0)<1$, a contradiction.

(2). Consider the corresponding marginal joint distributions
$V(i_1,  \dots, i_k)$ and $V(j_1,\dots, j_l)$.  Since
$|\Dmn(V(i_1, \dots, i_k))|<\infty$, and
$$\tau_{\alpha^P}^{\{V_{i_1}, \dots, V_{i_k}\}|V(j_1, \dots, j_l)}=1= \tau_{\alpha^P}^{V(j_1, \dots, j_l)|\{V_{i_1}, \dots,
V_{i_k}\}},$$ we have that $$|\Dmn(V(i_1, \dots, i_k))|=|\Dmn(V(j_1, \dots, j_l))|,$$ which is
obviously up bounded by
$$\min(\prod_{s=1}^km_{V_{i_s}},\prod_{t=1}^lm_{V_{j_t}}).$$

(3). There exists a variable, say,
 $V_{j_1}$, such that
  $$Ep(V_{j_1})=\min_{1\leq i\leq n}Ep(V_i).$$
  Let $V_{j_2}$ be a variable such that
  $$Ep((V_{j_1}, V_{j_2}))=\min_{i\neq j_1,1\leq i\leq n}Ep((V_{j_1},V_i)).$$
 Notice that for any $i\neq j_1$,
  \begin{align*}
   Ep((V_{j_1},V_i))
       &\leq \sum_{\eta,\beta}p(V_{j_1}=\eta,V_i=\beta)p(V_{j_1}=\eta)\\
       &= Ep(V_{j_1}),
  \end{align*}
  and the equality holds if and only if whenever
  $p(V_{j_1}=\eta, V_i=\beta)\neq 0$, we have
  $p(V_i=\beta |V_{j_1}=\eta)=1$; in other words, if and only if $V_i$ is completely dependent of $V_{j_1}$, the
  equality holds.  Thus
  $Ep((V_{j_1},V_{j_2}))\leq Ep(V_{j_1})$; and if the equality
  holds then for any $i\neq j_1$, $V_i$ is completely dependent of
  $V_{j_1}$; and in this case, $k=1$ then we are done.  Assume $Ep((V_{j_1},V_{j_2}))<
  Ep(V_{j_1})$.
  Then there exists
  $V_{j_3}\in\{V_1,\dots,V_n\}\setminus\{V_{j_1}, V_{j_2}\}$ such
  that
  $$Ep((V_{j_1}, V_{j_2}, V_{j_3}))=\min_{i\in\{1,\dots,n\}\setminus\{j_1,j_2\}}Ep((V_{j_1},V_{j_2},V_i)).$$
 Similarly we have $Ep((V_{j_1}, V_{j_2}, V_{j_3}))\leq Ep((V_{j_1},
 V_{j_2}))$.  If the equality holds, $k=2$; otherwise, we continue
 until for some $K<n$, $Ep((V_{j_1}, \dots, V_{j_{K+1}}))=
 Ep((V_{j_1},\dots, V_{j_K}))$.

Let $Y=(V_{j_1},\dots, V_{j_K})$, and
 $Y_{\hat{j}}=(V_{j_1},\dots, \hat{V_{j_h}},\dots,
 V_{j_K})$, where\\ $(V_{j_1},\dots, \hat{V_{j_h}},\dots,
 V_{j_K})$ stands for the joint distribution $Y$ with $V_{j_h}$ being removed.  Calculate $\tau_{\alpha^P}^{Y|Y_{\hat{h}}}$, for all $h=1,\dots,
 K$.  If for some $h_1$, $\tau_{\alpha^P}^{Y|Y_{\hat{h_1}}}=1$, then $V_{j_{h_1}}$ is redundant.
 Remove this redundant variable.  In this case, let
 \begin{equation*}
 Y_{\hat{h_1}, \hat{h}}=
  \begin{cases} &(V_{j_1},\dots,
  \hat{V_{j_h}},\dots,\hat{V_{j_{h_1}}}
 V_{j_K}), \mbox{ if }h<h_1,\\
 &(V_{j_1},\dots,
  \hat{V_{j_{h_1}}},\dots,\hat{V_{j_h}}
 V_{j_K}), \mbox{ if }h>h_1.
 \end{cases}
 \end{equation*}
 Calculate $\tau_{\alpha^P}^{Y|Y_{\hat{h_1}, \hat{h}}}$, for all
 $h\in\{1,\dots,\hat{h_1},\dots,
 K\}$.  If for some $h_2$, $\tau_{\alpha^P}^{Y|Y_{\hat{h_1},\hat{h_2}}}=1$, then $V_{j_{h_2}}$ is redundant.
 Remove this redundant variable.  Continue in this fashion to remove all possible redundant variables from
 $V_{j_1},\dots, V_{j_K}$.  Then we obtain a structural base for
 $S$.  In general, there can be more bases for $S$.  Choose a base containing the smallest number of variables.  Then
 this base is the desired one.
$\square$

\end{document}